\pdfoutput=1

\documentclass[11pt]{article}

\usepackage[preprint]{coling}

\usepackage{times}
\usepackage{latexsym}

\usepackage{natbib}

\usepackage[T1]{fontenc}

\usepackage[utf8]{inputenc}

\usepackage{microtype}

\usepackage{inconsolata}

\usepackage{graphicx}

\usepackage{booktabs}
\usepackage{placeins}
\usepackage{listings}
\usepackage{xcolor}
\usepackage{hyperref}
\usepackage{tabularx}  
\usepackage{array}     
\usepackage{url}

\title{Fineweb-Edu-Ar: Machine-translated Corpus \protect\\ to Support Arabic Small Language Models}

\author{
  Sultan Alrashed\thanks{Equal contribution. The order is random.} \\
  Saudi Data \& Artificial Intelligence Authority (SDAIA) \\
  \texttt{srashed@sdaia.gov.sa} \\
  \AND
  Dmitrii Khizbullin\footnotemark[1] \\
  King Abdullah University of Science \& \\ Technology (KAUST) \\
  \texttt{dmitrii.khizbullin@kaust.edu.sa} \\
  \And
  David R. Pugh \\
  KAUST \\
  \texttt{david.pugh@kaust.edu.sa}
}

\definecolor{punct}{RGB}{153,153,153}
\definecolor{delim}{RGB}{20,105,176}

\lstdefinelanguage{json}{
    basicstyle=\ttfamily\small,
    numbers=left,
    numberstyle=\scriptsize\color{gray},
    stepnumber=1,
    numbersep=8pt,
    showstringspaces=false,
    breaklines=true,
    frame=single,
    backgroundcolor=\color{white},
    literate=
     *{0}{{{\color{blue}0}}}{1}
      {1}{{{\color{blue}1}}}{1}
      {2}{{{\color{blue}2}}}{1}
      {3}{{{\color{blue}3}}}{1}
      {4}{{{\color{blue}4}}}{1}
      {5}{{{\color{blue}5}}}{1}
      {6}{{{\color{blue}6}}}{1}
      {7}{{{\color{blue}7}}}{1}
      {8}{{{\color{blue}8}}}{1}
      {9}{{{\color{blue}9}}}{1}
      {:}{{{\color{punct}{:}}}}{1}
      {,}{{{\color{punct}{,}}}}{1}
      {\{}{{{\color{delim}{\{}}}}{1}
      {\}}{{{\color{delim}{\}}}}}{1}
      {[}{{{\color{delim}{[}}}}{1}
      {]}{{{\color{delim}{]}}}}{1},
    keywordstyle=\color{black}\bfseries,
    stringstyle=\color{green!50!black},
    commentstyle=\color{gray},
    morestring=[b]",
    morestring=[d]'
}

\begin{document}

\maketitle


\begin{abstract}
As large language models (LLMs) grow and develop so too, do their data demands. This is especially true for multilingual LLMs, where the scarcity of high quality and readily available data online has led to a multitude of synthetic dataset generation approaches. A key technique seen in this space is machine translation (MT), where high quality English text is adapted to a target, comparatively low-resource, language. 

In this report we introduce \textit{FineWeb-Edu-Ar}, a machine-translated version of HuggingFace's exceedingly popular (deduplicated) FineWeb-Edu dataset. To the best of our knowledge FineWeb-Edu-Ar is the biggest publicly available machine-translated Arabic dataset out there, with its size of 202B tokens of an Arabic-trained tokenizer.

\makeatletter
\ifcoling@anonymize
The data is available on HuggingFace at [link].
\else
  The data is available on HuggingFace\footnote{\url{https://huggingface.co/datasets/kaust-generative-ai/fineweb-edu-ar}}.
\fi
\makeatother

\end{abstract}

\section{Introduction}
Natural Language Processing (NLP) has seen tremendous strides in recent years with the advent of LLMs, we have seen models scale up to more than 100 billion parameters \cite{NEURIPS2020_gpt3}. With the ever growing demand for scale, we have seen a substantial focus on large corpora to train these LLMs on \cite{hoffmann2022training}. Although data quality mattered, the focus had mostly been on data quantity.

More recently however, we see small language models (SLMs) trained on a much smaller quality-focused corpus \cite{gunasekar_textbooks_2023}, \cite{allal2024SmolLM}. These models exceed expectations for their size, outperforming certain much larger models on various benchmarks \cite{abdin2024phi3technicalreporthighly}. This gave rise to new potential use cases for language models on edge devices with compute constraints \cite{lu2024small}, \cite{Gunter2024AppleIF}. Many languages, including Arabic, suffer from a distinct lack of the same kind of high quality, educational focused, and readily available data that allowed other small language models to flourish.

\begin{figure}[t]
  \centering
  \includegraphics[width=1.0\linewidth]{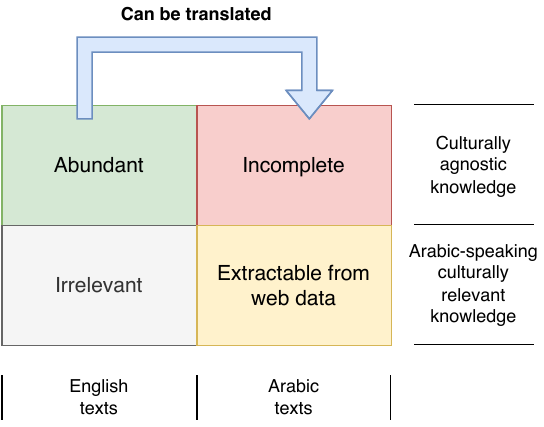}
  \caption{Quadrants of corpora availability.}
  \label{fig:translation_quadrants}
\end{figure}

\FloatBarrier

In order to illustrate the current situation with the training data, we've organized it in 4 quadrants: English and Arabic texts versus culturaly-agnostic and culturally-specific knowledge as in Figure \ref{fig:translation_quadrants}. While generic English data (left top quadrant) is abundant, and Arabic culturally specific data (right bottom quadrant) can be extracted from the web archives \cite{Aloui2024101BA}, Arabic texts with the generic knowledge (top right quadrant) are scarce. Fortunately, the latter can be reinforced with the machine translated texts sourced from generic English texts.

In an effort to support small language model pre-training, we publicly release \textit{FineWeb-Edu-Ar}, a machine translated version of \texttt{HuggingFaceTB/smollm-corpus} \cite{allal2024SmolLM}, the deduplicated version of Huggingface's FineWeb-Edu \cite{lozhkov2024fineweb-edu}. \texttt{HuggingFaceTB/smollm-corpus} that enabled their model, SmolLM, to achieve remarkable results for its size. We hope that this will support the field of Arabic SLMs. The dataset is released under CC-BY-NC-4.0 license.

Additionally, to support community efforts in English to Arabic machine translation, we provide a preliminary analysis into the performance of several translation models.

In general, we see two ways to obtain the educational-grade Arabic language dataset: (1) translating the English FineWeb-Edu, and (2) adapting the FineWeb-Edu pipeline for Arabic with subsequent re-running it on CommonCrawl. One notable attempt to do the latter is \cite{ArabicWeb24} which, however, does not feature semantic filtration, possibly due to the small total amount of Arabic content in the Web compared to English. Whereas \cite{ArabicWeb24} is a valuable dataset, specific to Arabic-speaking countries and obtained from the native speakers, we choose to follow the option of translation, since the source dataset (\texttt{FineWeb-Edu-dedup}) a lot of information that Arabic native speakers choose to get from English sources, and since these two ways are, for the most part, complimentary.

As additional evidence for the acute interest to small condensed-knowledge datasets of specific languages we would like to mention Chinese-FineWeb-Edu \cite{chinese_fineweb_edu} which is the FineWeb-Edu semantic filtration pipeline run for Chinese language.

Our contributions are:

\begin{enumerate}
    \item An open sourced machine translated version of HuggingFace's FineWeb-Edu to Arabic.
    \item An analysis into the performance and computational requirements of 12 models for English to Arabic machine translation.
\end{enumerate}

\section{Related Works}
\subsection{Small Language Models}

Although scaling LLMs to ever-increasing sizes \cite{touvron_llama_2023} has been the focus for a long time, we are seeing an increasing focus on training SLMs.

Unlike their larger counterparts, these SLMs benefit a lot more from the quality of the corpus they are trained on rather than the quantity. Works such as Huggingface's SmolLM \cite{allal2024SmolLM} show this in their benchmark scores \ref{smollm_benchmark}, where Qwen2-1.5B, a model trained on 7 trillion tokens \cite{yang2024qwen2technicalreport}, scores lower than SmolLM-1.7B, which was only trained with 1 trillion tokens.

\begin{table}[ht]
  \centering
  \begin{tabular}{lll}
    \hline
    \textbf{Model Name} & \textbf{MMLU} & \textbf{ARC} \\
    \hline
    \texttt{Qwen1.5-1.8B} & 33.46 & 47.08 \\
    \texttt{Qwen2-1.5B} & 37.87 & 48.20 \\
    \textbf{\texttt{SmolLM-1.7B}} & \textbf{39.97} & \textbf{61.55} \\
    \hline
  \end{tabular}
  \caption{\label{smollm_benchmark}
    Benchmark scores comparing <2b parameter language models, taken from SmolLM 
  }
\end{table}

This disparity in training size and benchmark scores further cements the need for high quality data for SLMs.

\subsection{Arabic Machine Translation}

In the CommonCrawl dataset, one of the largest web-crawled datasets, Arabic content is over two orders of magnitude less common than English content \cite{wenzek2019ccnet}. This discrepancy has naturally led to a need for more Arabic data, machine translation allows us to fill that necessary gap.

One of the largest efforts towards Arabic machine translation has been the work on Aya Dataset \cite{singh2024aya}, which contains over 6 million entries of Modern Standard Arabic (MSA) data translated from English sources. Taking inspiration from their successful usage of \texttt{nllb-200-3.3B} helped further inform our decision on using \texttt{nllb-200-distilled-600M}. While Aya Dataset focused on instructional tuning data, we had shifted our focus to a gap in English to MSA machine translated pretraining corpora. To match the scale necessary for pretraining data, we translate approximately 190 million rows, as seen in table \ref{tab:data_properties}.

\section{MT Model Assessment}
\subsection{Models Chosen for Assessment}

To find out which model would be ideal to translate such a large corpus, we conduct an investigation into the translation and run time performance of a collection of popular English to Arabic encoder-decoder transformers alongside multilingual decoder-only transformers.

For our encoder-decoder transformers, we selected from a set of popular machine translation models. We went with the 600m and 1.3b variants of Facebook's NLLB \cite{costa2022no} and the large English to Arabic version of the University of Helsinki's OPUS model \cite{tiedemann-thottingal-2020-opus}.

For our decoder-only transformers, we selected from a collection of generally popular models and ones that excel in Arabic. We went with the 8b variant of Aya-23 8b \cite{Aryabumi2024Aya2O}, the 8b variant of Llama 3.1 \cite{dubey2024llama3herdmodels}, the 3b variant of Llama 3.2 \cite{meta2024llama32}, both the 7b and 32b variants of Qwen 2.5 \cite{yang2024qwen2}, Mistral Small Instruct \cite{mistral2024models}, Command-R \cite{command-r}, and Tower-Instruct \cite{tower_llm_2024}.

\subsection{MT Model Quality Assessment}

\begin{figure*}[!htbp]
  \centering
  \includegraphics[width=1.0\linewidth]{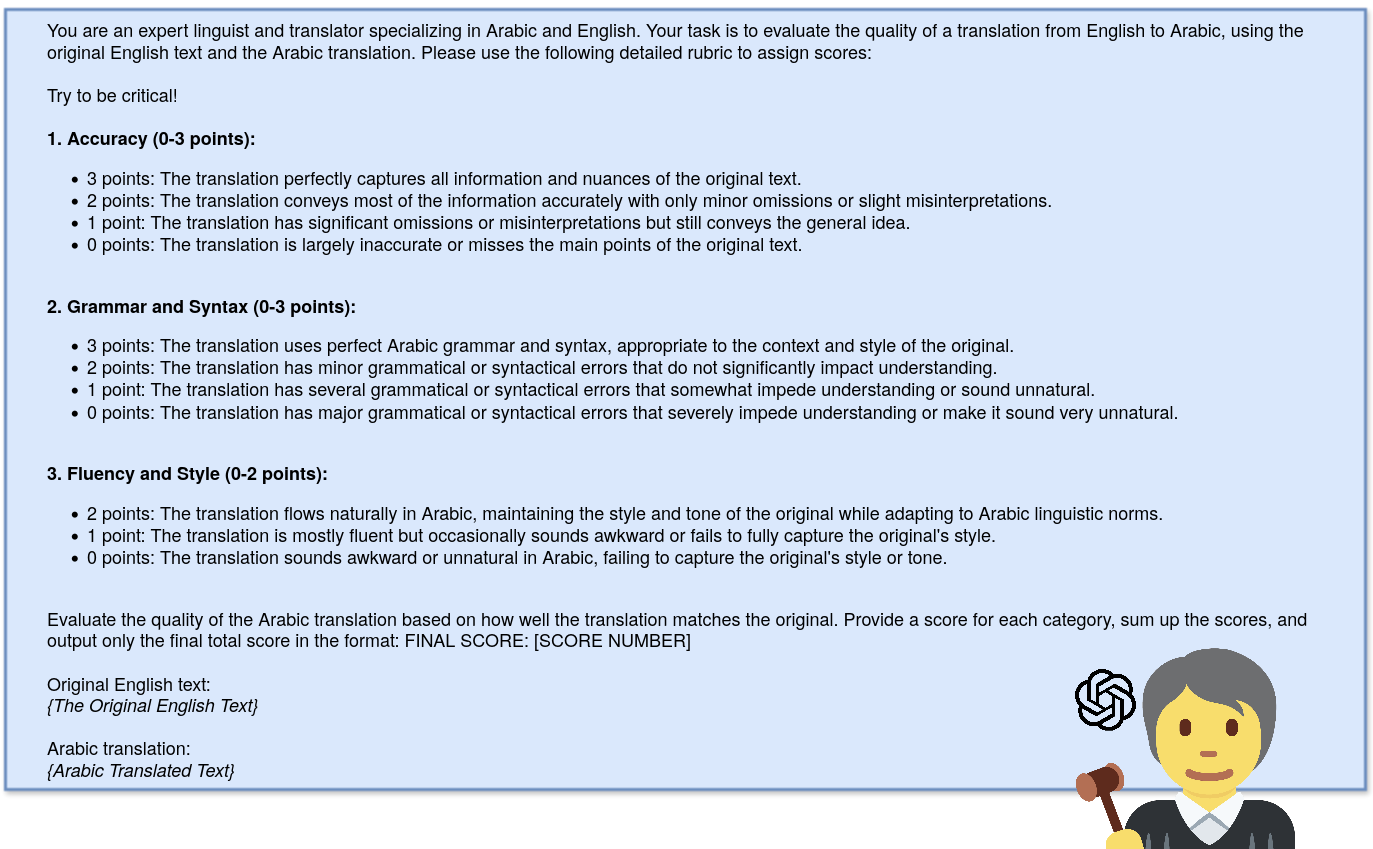}
  \caption{The prompt that we've designed for GPT-4o to evaluate the translation quality of a given text.}
  \label{fig:gpt4o_judge_prompt}
\end{figure*}

To assess MT quality in an automatic and reproducible manner, we employ the LLM-as-a-Judge approach \cite{zheng2024judging}, \cite{fu2023gptscore}. We prompt GPT-4o to assess the quality of translation from 3 different angles: accuracy (0-3 points), grammar and syntax (0-3 points), alongside fluency and style (0-2 points). We provide both original English and the translated Arabic passages as a part of the prompt. See Figure \ref{fig:gpt4o_judge_prompt} for the exact prompt that we've designed. In order to make our micro-benchmark for MT models dataset-independent, we prepare a set of 3 passages to run the translation on. The passages that compose our micro-benchmark are listed in Appendix~\ref{sec:micro_benchmark}. To get a more stable estimate of the quality, we re-run the translation 10 times with 0.1 temperature, average the scores and round them. We add up the scores for all passages to get the final quality score between 0 and 24 for every MT model. 
Further, we perform runtime performance benchmarking for all models on a single A100 80GB GPU
with \texttt{flash\_attention\_2} enabled for whichever model supports it. 

\begin{figure*}[!htbp]
  \centering
  \includegraphics[width=0.98\linewidth]{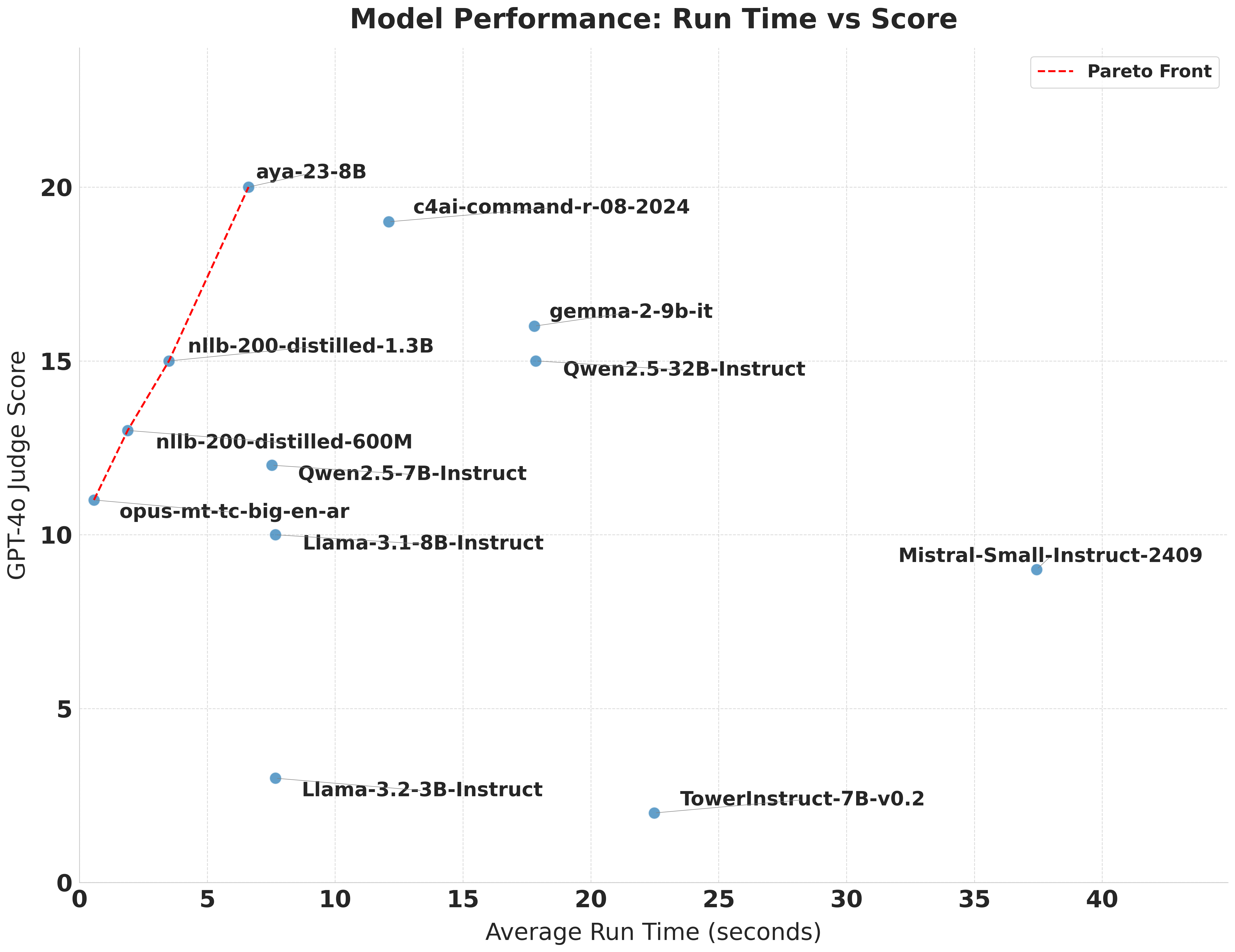}
  \caption{Scatter plot of gpt-4o judge score against each model's runtime.
  \textit{nllb-200-distilled-600m} lies on the Pareto frontier shown with the red dashed line.
  }
  \label{fig:experiments}
\end{figure*}

We build a map of all evaluated models in score vs. runtime coordinates, see Figure \ref{fig:experiments}. We find 4 models that lie on the Pareto frontier.

We choose \texttt{nllb-200-distilled-600M} as the best MT model within the computational budget of 500 A100 GPU-days.

\begin{figure*}[!htbp]
  \centering
  \includegraphics[width=0.75\linewidth]{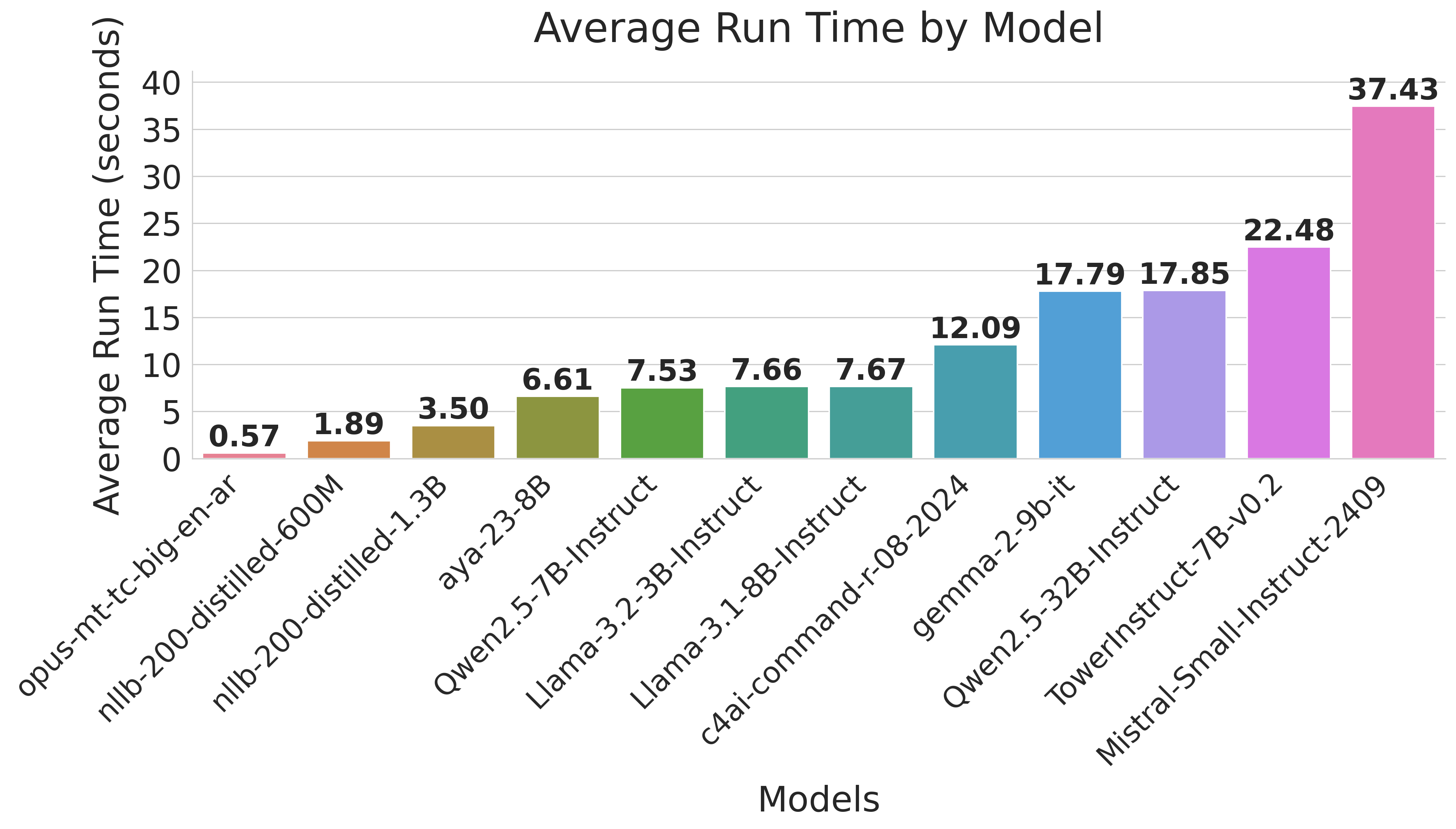} \\
  \vspace{0.5cm}
  \includegraphics[width=0.75\linewidth]{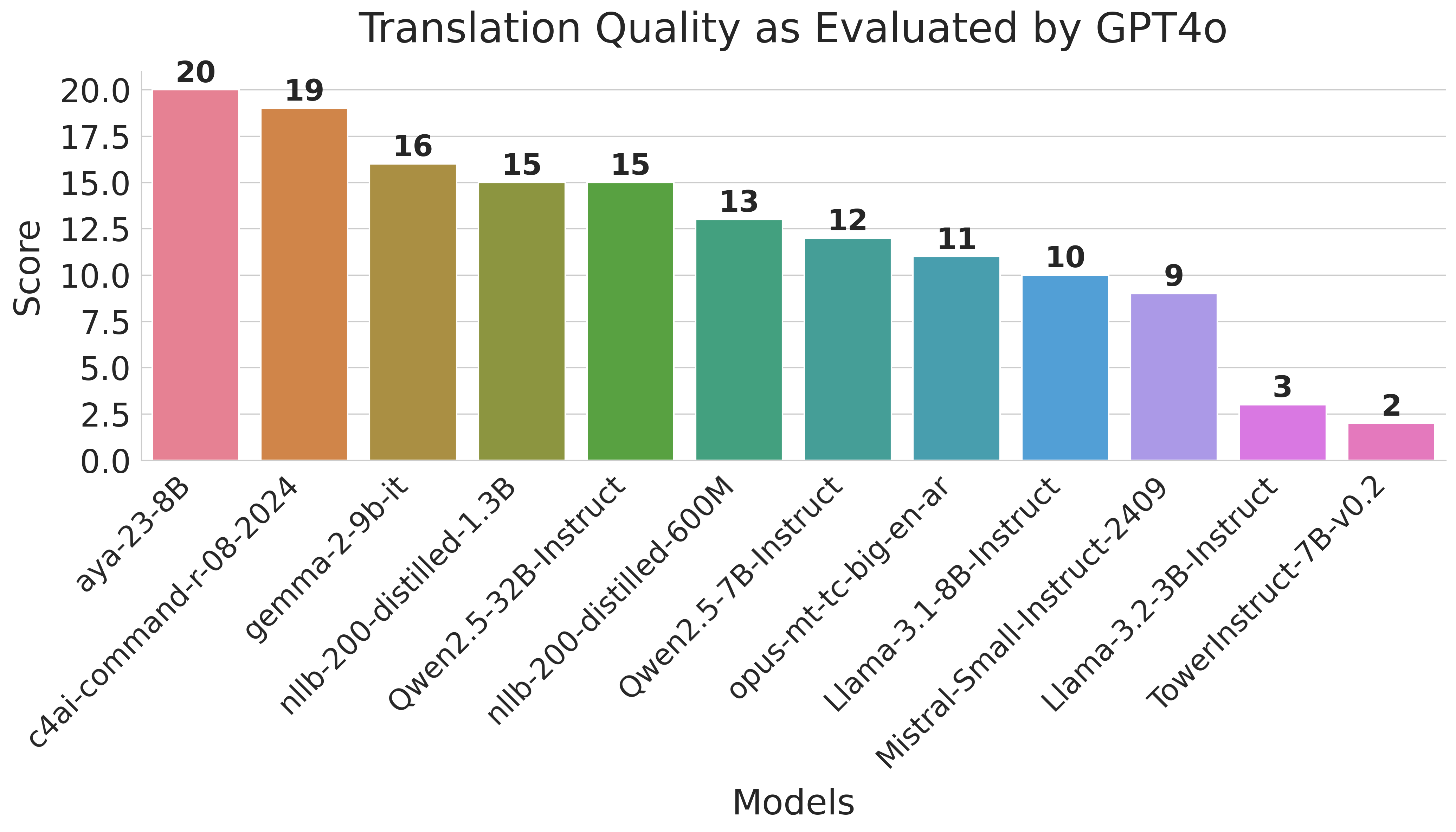}
  \caption{Models' runtime and score evaluated on our micro-benchmark. The maximal score that a translation model can achieve is 24.}
  \label{fig:model_evaluation}
\end{figure*}

\section{Dataset}

\begin{table*}[htbp]
\centering
\begin{tabular}{ll}
\toprule
\textbf{Parameter} & \textbf{Value} \\ \midrule
Translation model & \texttt{facebook/nllb-200-distilled-600M} \\
Total passages & 189,405,457 (189M) \\
Number of shards, each language & 1813 \\
Total jsonl size, Ar & 937 GB \\
Total jsonl size, En & 844 GB \\
Total zip size, Ar & 267 GB \\
Total zip size, En & 311 GB \\
GPT-2 tokens, Ar & 551,094,680,322 (551B) \\
GPT-2 tokens, En & 192,121,988,036 (192B) \\
NLLB tokenizer tokens, Ar & 202,379,768,558 (202B) \\
NLLB tokenizer tokens, En & 210,765,476,860 (210B) \\
GPU resources & 20 days of 24 A100 GPUs (480 GPU-days) \\
\bottomrule
\end{tabular}
\caption{Resource usage and details for the translation model.}
\label{tab:data_properties}
\end{table*}

\makeatletter
\ifcoling@anonymize
The data is available on HuggingFace at [link]. The code used to create the dataset is available at [link].
\else
  The data is available on HuggingFace\footnote{\url{https://huggingface.co/datasets/kaust-generative-ai/fineweb-edu-ar}}. The code used to create the dataset is available on GitHub\footnote{\url{https://github.com/SulRash/llm-translation-pipeline}}.
\fi
\makeatother

\subsection{Generation details}

Due to the limited context window of NLLB, there were multiple approaches we could have taken to split up our data for translation:

\begin{itemize}
    \item Sliding window with overlap: The input text is split into 200 token chunks, with a 50 token overlap.
    \item Sliding window with no overlap: The input text is split into 200 token chunks.
    \item Sliding window split by sentence: The input text is split into the maximum number of sentences present in the 200 tokens chunk. If the sentence is longer than 200 tokens, we take the whole 200 tokens.
\end{itemize}

We resorted to a sliding window approach with no overlap to align with our goals of containing costs and CO$_2$ emissions. In particular, \texttt{flash\_attention\_2} is sensitive to the use of padding tokens \cite{kundu2024enhancingtrainingefficiencyusing}, which eliminates option 3.
Each text is broken up into 200 token chunks and translated separately as part of a larger batch, that is later concatenated.

Furthermore, to reduce redundant processing and conserve emissions, we processed only the deduplicated version of Fineweb-Edu present in the training set of SmolLM, a successful English SLM.

\subsection{Properties}

Dataset properties are summarized in Table \ref{tab:data_properties}. Dataset schema is defined in Listing~\ref{lst:json_schema}. The dataset contains both the Arabic translated and the original English passages available at corresponding indices.

\section{Discussion and Limitations}
We choose the translation model among those lying on the Pareto frontier and at the same time satisfying the computational budget. Within these constraints the produced dataset strikes a balance between the translation quality and cost. Still, the question of how good the data is for creation of foundational models remains open, and we leave it to the community to find out.

The proposed dataset can be considered noisy in the sense of translation inaccuracies. Furthermore, the knowledge domain of FineWeb-Edu is one of English speaking countries and may not include enough regional facts of the Arabic-speaking countries. Nevertheless, we argue that the sheer amount of language-agnostic knowledge that FineWeb-Edu contains may be valuable for pre-training SLMs.

A thorough analysis of the alignment of LLM-as-a-Judge with independent Arabic native speakers' preferences is left out of the scope of this work and is suggested for the future study.

\subsection{CO2 Emissions}

Using the OECD's 2014 yearly average carbon efficiency of 0.432 kgCO$_2$eq/kWh, alongside the 400W TDP of the GPU used, the full translation total emissions are estimated to be 1990 kgCO$_2$eq for the 11520 GPU-hours. These estimations were conducted using the MachineLearning Impact calculator\footnote{\texttt{https://mlco2.github.io/impact\#compute}} presented in \cite{lacoste2019quantifying}.

Using the runtime numbers present in figure \ref{fig:experiments}, we can estimate the equivalent emission costs of using the other models we heavily considered in table \ref{tab:emissions}.

\begin{table}[h]
  \centering
  \begin{tabular}{ll}
    \hline
    \textbf{Model Name} & \textbf{Emissions,} \\
    & \textbf{kgCO$_2$eq} \\
    \hline
    \texttt{aya-23-8B} & 6975 \\
    \texttt{nllb-200-distilled-1.3B} & 3693 \\
    \textbf{\texttt{nllb-200-distilled-600M}} & \textbf{1990} \\
    \hline
  \end{tabular}
  \caption{\label{tab:emissions}
    Comparing the emissions of models most viable for translation.
  }
\end{table}

The disparity in emissions is significant. As part of promoting training SLMs, we also want to promote climate awareness and consciousness in this space. This drove our decision towards \texttt{nllb-200-distilled-600M}.

\section{Conclusion}

FineWeb-Edu-Ar is the biggest Arabic machine-translated dataset that is openly available for non-commercial purposes. The proposed dataset took around 11520 A100 GPU-hours to produce at the cost of \$46,000; and we suggest that it is of significant value to starter academic teams who do not have resources to generate their own synthetic data at this scale.


Particularly, we hope this dataset allows the community to explore the space of Arabic small language models.

\FloatBarrier

\makeatletter
\ifcoling@anonymize
\else
  \section{Acknowledgements}

  This work was supported by the SDAIA-KAUST Center of Excellence in Data Science and Artificial Intelligence (SDAIA-KAUST AI).
\fi
\makeatother

\bibliography{citations}

\FloatBarrier

\appendix

\section{JSON data schema}

\begin{figure}[htbp]
\centering
\begin{lstlisting}[language=json, caption={JSON Schema for the data.}, label={lst:json_schema}]
{
  "$schema": "http://json-schema.org/draft-07/schema#",
  "title": "FineWebArSchema",
  "type": "object",
  "properties": {
    "text": {
      "type": "string"
    }
  },
  "required": ["text"],
  "additionalProperties": false
}
\end{lstlisting}
\end{figure}

\section{Examples of passages}

Examples of passages are displayed in Table \ref{tab:examples}.

\begin{table*}[!ht]
\centering
\begin{tabularx}{\textwidth}{|X|X|X|}
\hline
\textbf{Original English} & \textbf{Translated Arabic} & \textbf{Translated back to English by Google Translate} \\ \hline
\includegraphics[width=0.99\linewidth]{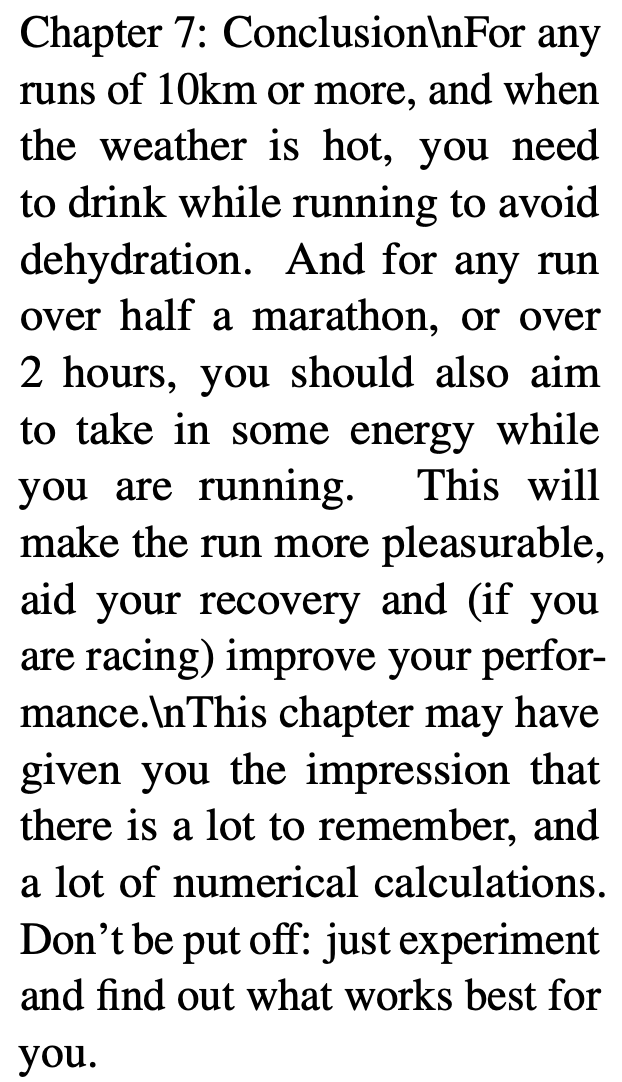}
&
\includegraphics[width=\linewidth]{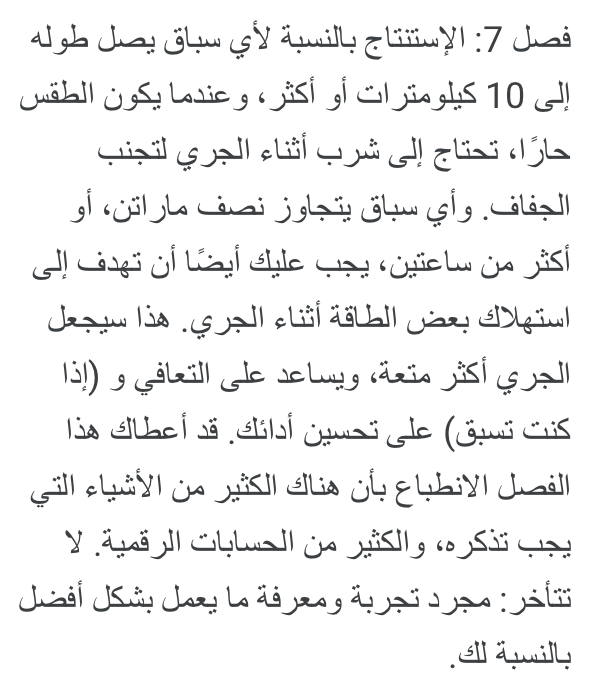}
&
\includegraphics[width=0.99\linewidth]{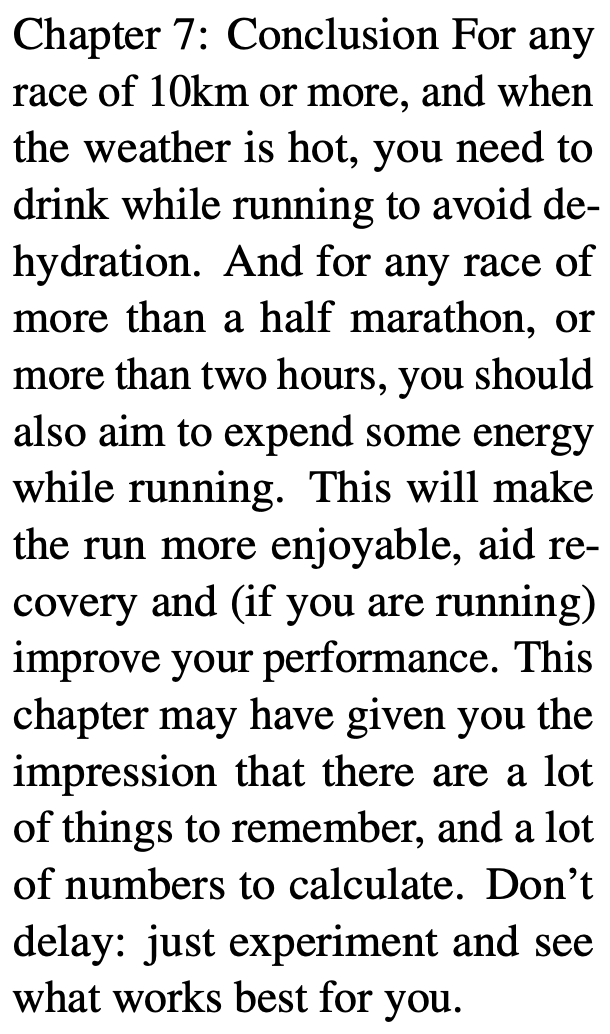}
\\ \hline
\end{tabularx}
\caption{An example of a passage from the dataset.
}
\label{tab:examples}
\end{table*}

\section{Texts in the MT model evaluation micro-benchmark}
\label{sec:micro_benchmark}
\nopagebreak[4]

See Listing \ref{lst:micro_benchmark}.

\begin{figure*}[!htbp]
\centering
\begin{lstlisting}[language=json, caption={Micro-benchmark contents.}, label={lst:micro_benchmark}]
{
    "AR_WIKI_TEXT_TO_TRANSLATE": "Arabic is the third most widespread official language after English and French, one of six official languages of the United Nations, and the liturgical language of Islam. Arabic is widely taught in schools and universities around the world and is used to varying degrees in workplaces, governments and the media. During the Middle Ages, Arabic was a major vehicle of culture and learning, especially in science, mathematics and philosophy. As a result, many European languages have borrowed words from it. Arabic influence, mainly in vocabulary, is seen in European languages (mainly Spanish and to a lesser extent Portuguese, Catalan, and Sicilian) owing to the proximity of Europe and the long-lasting Arabic cultural and linguistic presence, mainly in Southern Iberia, during the Al-Andalus era. Maltese is a Semitic language developed from a dialect of Arabic and written in the Latin alphabet. The Balkan languages, including Greek and Bulgarian, have also acquired many words of Arabic origin, mainly through direct contact with Ottoman Turkish.",
    "CLOCK_BLOG_TEXT_TO_TRANSLATE": "I thought about making something similar, but decided that a high-tech solution isn't always the right one. Instead I made a clock, or rather a clock face. My rationale was as follows: the bin days don't change frequently so no need to call an API to get them, and anyway most councils don't have an API for this sort of thing. Also, I really didn't want yet another thing with a wall wart, or WiFi to configure, or code to debug (there comes a time in every programmer's life when they can't face debugging yet another thing that should be simple and just work). \nNot everything needs to be Turing Complete! But you can buy cheap clock mechanisms where the hands go full circle in seven days instead of 12 hours.",
    "AI_REUTERS_TEXT_TO_TRANSLATE": "STOCKHOLM/SAN FRANCISCO, Sept 28 (Reuters) - In the early years, getting AI models like ChatGPT or its rival Cohere to spit out human-like responses required vast teams of low-cost workers helping models distinguish basic facts such as if an image was of a car or a carrot.\nBut more sophisticated updates to AI models in the fiercely competitive arena are now demanding a rapidly expanding network of human trainers who have specialized knowledge -- from historians to scientists, some with doctorate degrees. 'A year ago, we could get away with hiring undergraduates, to just generally teach AI on how to improve,' said Cohere co-founder Ivan Zhang, talking about its internal human trainers."
}
\end{lstlisting}
\end{figure*}

\end{document}